\documentclass[lettersize,journal]{IEEEtran}

\usepackage{amsthm}
\usepackage{amssymb}

\usepackage{adjustbox}  
\usepackage[ruled,vlined,linesnumbered]{algorithm2e}
\SetAlFnt{\small}
\SetAlCapFnt{\small}
\SetAlCapNameFnt{\small}
\SetArgSty{textnormal}
\SetInd{0.1em}{0.5em}
\SetCommentSty{emph}
\SetAlgoSkip{}
\usepackage{etoolbox}
\makeatletter
\patchcmd{\@algocf@start}
  {-1.5em}
  {0pt}
  {}{}
\makeatother

\usepackage{enumitem}
\usepackage{hyperref}
\usepackage{tabularx} 
\usepackage{booktabs}   
\usepackage{array}      
\usepackage{siunitx}
\usepackage{amsmath}
\usepackage{xcolor} 
\usepackage{multirow}
\usepackage{graphicx} 
\usepackage{siunitx}

\usepackage{caption}

\definecolor{RED}{rgb}{0.933, 0.49, 0.42}  
\definecolor{CYAN}{rgb}{0.42, 0.78, 0.95}

\usepackage{soul}
\usepackage{xcolor}
\sethlcolor{yellow!30}

\soulregister\ref7
\soulregister\cref7
\soulregister\Cref7
\soulregister\eqref7
\soulregister\cite7
\soulregister\citep7
\soulregister\citet7

\newcommand{\jiaqi}[1]{{#1}}

\newcommand{\rTwoR}{\text{R2R}}
\newcommand{\sTwoW}{\text{S2W}}
\newcommand{\dAndE}{\text{DnE}}
\newcommand{\switch}{DS} 
\newcommand{\trajReplan}{GTR} 
\newcommand{\singleAgentTrajReplan}{STR} 
\newcommand{\framework}{CREST} 

\DeclareMathOperator*{\argmin}{arg\,min}

\title{CREST: Constraint-Release Execution\\for Multi-Robot Warehouse Shelf Rearrangement}

\author{Jiaqi Tan, Yudong Luo, Sophia Huang, Yifan Yang, and Hang Ma%
\thanks{Manuscript received: November 3, 2025; Revised: January 23, 2026; Accepted: March 8, 2026. This paper was recommended for publication by Editor Chao-Bo Yan upon evaluation of the Associate Editor and Reviewers' comments. This work was supported by the NSERC under grant number
RGPIN2020-06540 and a CFI JELF award.}%
\thanks{Jiaqi Tan, Yifan Yang, and Hang Ma are with the School of Computing Science, Simon Fraser University, Burnaby, BC V5A 1S6, Canada.}%
\thanks{Yudong Luo is with HEC Montr\'eal, Montr\'eal, QC H3T 2A7, Canada.}%
\thanks{Sophia Huang is with Purdue University, West Lafayette, IN 47907 USA.}%
\thanks{Corresponding author: Jiaqi Tan (e-mail: jiaqit@sfu.ca).}%
\thanks{Digital Object Identifier (DOI): see top of this page.}%
}

\begin{document}

\markboth{IEEE ROBOTICS AND AUTOMATION LETTERS. PREPRINT VERSION. ACCEPTED MARCH, 2026}%
{Tan et al.: CREST: Constraint-Release Execution for Multi-Robot Warehouse Shelf Rearrangement}


\maketitle


\begin{abstract}
Double-Deck Multi-Agent Pickup and Delivery (DD-MAPD) models the multi-robot shelf rearrangement problem in automated warehouses. MAPF-DECOMP is a recent framework that first computes collision-free shelf trajectories with a MAPF solver and then assigns agents to execute them. While efficient, it enforces strict trajectory dependencies, often leading to poor execution quality due to idle agents and unnecessary shelf switching. We introduce CREST, a new execution framework that achieves more continuous shelf carrying by proactively releasing trajectory constraints during execution. Experiments on diverse warehouse layouts show that CREST consistently outperforms MAPF-DECOMP, reducing metrics related to agent travel, makespan, and shelf switching by up to 40.5\%, 33.3\%, and 44.4\%, respectively, with even greater benefits under lift/place overhead. These results underscore the importance of execution-aware constraint release for scalable warehouse rearrangement. Code and data are available at \url{https://github.com/ChristinaTan0704/CREST}.
\end{abstract}
\IEEEpubidadjcol

\section{Introduction}

\begin{figure}[t!]
\centering
\includegraphics[width=0.75\columnwidth]{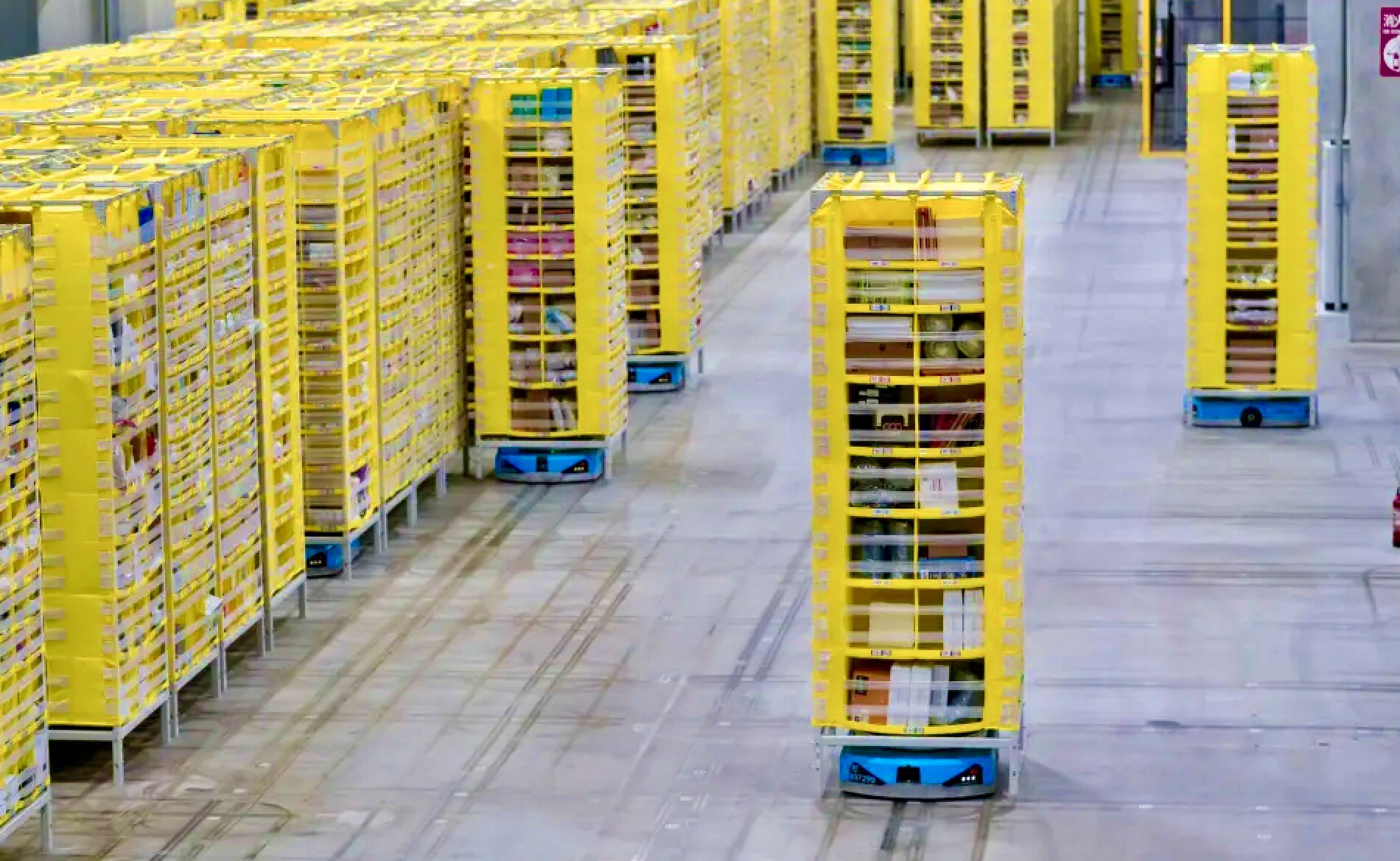}

\caption{Illustration of warehouse robots rearranging shelves~\cite{nypost2025robots}.}
\label{fig:fulfillment}
\vspace{-10pt}
\end{figure}

Automated warehouses rely on teams of mobile robots to transport and rearrange inventory shelves. A key challenge is to coordinate these agents so that shelves can be relocated safely without collisions, while maintaining high system throughput.
This problem is modeled as Double-Deck Multi-Agent Pickup and Delivery (DD-MAPD)~\cite{li2023double}, which generalizes both the well-studied Multi-Agent Path Finding (MAPF)~\cite{stern2019multi} and Multi-Agent Pickup and Delivery (MAPD)~\cite{ma2017lifelong} by allowing shelves to be lifted, carried, and placed arbitrarily, \jiaqi{thereby requiring collision avoidance at two levels: agent-agent and agent-obstacle interactions on the floor, and shelf-shelf and shelf-obstacle interactions during carrying.} This setting commonly arises in automated fulfillment centers (Fig.~\ref{fig:fulfillment})~\cite{wurman2008coordinating}.

A recent framework, MAPF-DECOMP~\cite{li2023double}, tackles DD-MAPD through a two-level decomposition. 
\jiaqi{First, a MAPF solver computes collision-free trajectories for all shelves. These trajectories are then partitioned into trajectory segments at shared locations. Second, a dependency graph is constructed over the segments based on the traversal order induced by the MAPF plan. Each trajectory segment is treated as a task and assigned to agents, analogous to MAPD with precedence constraints. This formulation is highly efficient, leveraging state-of-the-art MAPF solvers to scale to hundreds of shelves and agents within minutes.}

However, MAPF-DECOMP enforces these dependencies too strictly during trajectory execution. It inherits the online MAPD paradigm, where tasks arrive dynamically and assignments are repeated whenever agents complete their current tasks. 
In DD-MAPD, by contrast, all tasks and dependencies are known once shelf trajectories are computed. 
The execution phase thus resembles an offline scheduling problem, where the availability time of each agent can be determined once its current path is fixed, and the release time of a constrained shelf segment can be inferred once the constraining segments are planned.
MAPF-DECOMP does not fully exploit this global information, often leading to poor execution quality, such as agent idling, premature shelf switching, or missed opportunities for continued carrying. At the opposite extreme, fully coupled offline formulations that jointly optimize trajectories, assignments, and schedules~\cite{bachor2023multi,sherma2025agent} exploit this information but are computationally infeasible at warehouse scale.

\jiaqi{In this work, we strike a balance between conservative online execution and intractable offline optimization.} We introduce \textbf{CREST} (Constraint-Release Execution of Shelf Trajectories), a novel framework that retains the decomposition of MAPF-DECOMP while significantly improving the execution of shelf trajectories through lightweight, informed enhancements. By proactively releasing constraints, CREST enables shelves to progress more continuously without requiring a global rescheduling of all tasks. It integrates three complementary strategies that further reduce agent idling and unnecessary task switching: single trajectory replanning, dependency switching, and group trajectory replanning. These strategies leverage accurate execution information, such as actual agent and shelf available times and path costs, which becomes available only after assignments and local paths are computed. A purely offline planner that must decide all assignments and schedules in advance cannot access such information without resolving the full coupled problem.

We evaluate CREST on diverse warehouse scenarios, showing consistent improvements over MAPF-DECOMP in agent travel, makespan, and shelf switching, with even larger gains when lift/place overhead is considered. 
These results highlight the importance of execution-aware constraint release for scalable and efficient warehouse rearrangement.


\section{Related Work}

We survey related works on DD-MAPD and its algorithms.

\noindent\textbf{DD-MAPD and MAPF-DECOMP:} DD-MAPD extends prior research on multi-agent coordination by explicitly modeling movable objects transported by agents, capturing inter-shelf dependencies and two coupled levels of collision avoidance in warehouse operations. Equivalent formulations have appeared under different names, including Multi-Agent Transportation~\cite{bachor2023multi}, Multi-Agent and Multi-Rack Path Finding~\cite{makino2024marpf}, and Multi-Agent Warehouse Rearrangement~\cite{sherma2025agent}. Computing minimum-makespan DD-MAPD solutions is NP-hard~\cite{li2023double,bachor2023multi}. Fully coupled methods~\cite{bachor2023multi,makino2024marpf,sherma2025agent} that jointly optimize shelf trajectories, task assignments, and agent paths have been evaluated only on small instances (up to 8 agents and 16 shelves) with low success rates. MAPF-DECOMP~\cite{li2023double} remains the most scalable framework for DD-MAPD, demonstrating that decomposing shelf-trajectory planning from execution enables large-scale coordination.

\noindent\textbf{MAPF:}
MAPF focuses on computing collision-free paths for multiple agents or shelves. Representative solvers include methods based on CBS~\cite{sharon2015conflict,li2021pairwise,li2021eecbs}, prioritized planning~\cite{erdmann1987multiple,ma2019searching}, traffic rules~\cite{okumura2023improving}, and local search~\cite{li2022mapf}. They typically rely on Space-Time A*~\cite{silver2005cooperative} or SIPP~\cite{phillips2011sipp}, with extensions such as multi-label search~\cite{grenouilleau2019multi} for visiting multiple goals. MAPD~\cite{ma2017lifelong} extends MAPF to online pickup-and-delivery by repeatedly assigning tasks and replanning paths.


\noindent\textbf{Plan Execution and Dependency Graphs:}
MAPF plan execution studies how discrete, synchronized plans are executed asynchronously on physical robots. Temporal Plan Graphs (TPGs)~\cite{honig2016multi,ma2017multi} encode temporal dependencies as directed acyclic graphs, with recent extensions enabling dependency reversals to alter traversal order~\cite{berndt2020feedback,su2024bidirectional,JiangLinLi_STPGspeedup_2024}. These approaches primarily address small timing deviations due to robot dynamics or sensing uncertainty while largely preserving plan structure. In contrast, shelf-trajectory execution in MAPF-DECOMP exhibits much larger deviations: shelf trajectories are MAPF plans for movable objects whose execution depends on agent assignments and travel times, leading to both temporal and structural changes induced by inter-shelf dependencies. Similar TPG-based dependency reasoning has also been applied in coordinated manipulation and assembly~\cite{li2023fabrica,sun2024apexmr}.

In summary, prior research has addressed the scalability of DD-MAPD through decomposition and the robustness of trajectory execution through temporal plan representations. CREST bridges these two directions by retaining the decomposition structure of MAPF-DECOMP while introducing execution-aware constraint-release strategies. These strategies extend ideas from MAPF plan execution to handle timing variability arising from task assignment, agent paths, and inter-shelf dependencies during trajectory execution.

\section{Problem Definitions}

We now formalize DD-MAPD and its core definitions.

\subsection{DD-MAPD}

A DD-MAPD instance consists of $N$ agents $a_1,\dots,a_N$, $M$ shelves $s_1,\dots,s_M$, and a connected undirected graph $G=(V,E)$. Vertices $V$ represent locations and edges $E$ represent traversable connections. We consider non-trivial cases where $N\leq M$. \jiaqi{When $N > M$, at least one agent is always available to immediately execute any released shelf trajectory, reducing the problem to standard MAPF \cite{felner2026blue}.}

\noindent\textbf{Agents and Shelves:} Let $\pi_{a}(t)$ and $\phi_{s}(t)$ be  the locations of agent $a$ and shelf $s$ at discrete time step $t$.
Each agent $a_i$ starts at a distinct initial location $\pi_{a}(0)$ and either waits or moves to an adjacent location at each time step. Each shelf $s_j$ starts at a distinct pickup location $p_s = \phi_{s}(0)$ at $t=0$ and must be relocated to a distinct delivery location $d_s$ (if relocation is unnecessary, $p_s=d_s$). 
\jiaqi{By convention~\cite{stern2019multi}, the \textit{end time} of an agent path $\pi_a$ is the time step when the agent first arrives at its last location. For any $t$ beyond this time, the agent is assumed to wait at that location forever. Hence, $|\pi_a|$ equals the end time. The same convention applies to shelf trajectories.}
An agent can move beneath a shelf when it is not carrying, \textit{lift} it from below when co-located, carry it to a new location, and \textit{place} it down. \jiaqi{A shelf is marked as \textit{completed} once it reaches its delivery location, and \textit{uncompleted} otherwise.}

\noindent\textbf{Collision Avoidance:} Collisions must be avoided at both the agent (low) and shelf (high) levels. A vertex collision between agents $a$ and $a'$ occurs if $\pi_{a}(t) = \pi_{a'}(t)$; an edge collision occurs if $\pi_{a}(t)=\pi_{a'}(t+1)$ and $\pi_{a}(t+1)=\pi_{a'}(t)$. Analogous conditions apply to shelves using $\phi_s(t)$.

\noindent\textbf{Objective:} The objective is to compute collision-free paths for all agents to transport shelves from their pickup to delivery locations. Solution quality is often measured by the makespan $\max_{i=1,\ldots,N} |\pi_{a_i}|$ and the sum of costs $\sum_{i=1}^{N} |\pi_{a_i}|$. 
We evaluate two timing settings: (1) without lift/place overhead, where lift and place actions take no time; and (2) with overhead, where each such action takes \jiaqi{$\Delta \in \mathbb{N}$} time steps.

\subsection{Decomposition and Dependency Graph}

CREST retains the two-stage decomposition of MAPF-DECOMP~\cite{li2023double}: 
(1) compute safe 1-robust shelf trajectories via a MAPF solver, i.e., trajectories that remain collision-free under a one-timestep delay, 
and (2) assign agents to execute these trajectories by solving a MAPD instance with dependencies. The first stage thus produces a set of shelf trajectories that determine all inter-shelf dependencies. 

\noindent\textbf{Shelf Trajectory \& Plan, Safeness, and 1-Robustness:} A trajectory $\tau_s$ for shelf $s$ is a sequence of locations, called \textit{waypoints}, where the first waypoint $\tau_s(0)=\phi_s(0)$, the last waypoint $\tau_s(|\tau_s|-1)=d_s$, and each consecutive pair satisfies either $\tau_s(k+1)=\tau_s(k)$ (wait) or $(\tau_s(k),\tau_s(k+1))\in E$ (move).
A MAPF shelf plan $\mathcal{T} = \{\tau_s\}$ consists of trajectories without collisions, i.e.,  for all $s\neq s'$ and $k$, $\tau_s(k)\neq\tau_{s'}(k)$ and $(\tau_s(k), \tau_{s}(k+1))\neq (\tau_{s'}(k+1), \tau_{s'}(k))$.
The plan is \textit{safe} if \jiaqi{trajectories never pass through the agents’ initial locations, i.e. } for all $s$, $k$, and agents $a$, $\tau_s(k)\neq \pi_a(0)$, and \textit{1-robust} if 
\jiaqi{trajectories are guaranteed to remain collision-free under a one-timestep execution delay, i.e. } for all $s\neq s'$ and $k$, $\tau_s(k+1)\neq\tau_{s'}(k)$~\cite{li2023double}.

\noindent\textbf{Dependency Graph:} The dependencies in a MAPF shelf plan $\mathcal{T}$ are represented by a dependency graph $D$ following \cite{ma2017multi}. Each node $\tau_s(k)$ represents the event ``shelf $s$ at its $k$-th waypoint $\tau_s(k)$''. Two types of arcs define precedence: Type-1 (intra-shelf) $\tau_s(k) \rightarrow \tau_s(k+1)$ for all $s$ and $k=0,\ldots,|\tau_s|-2$ and Type-2 (inter-shelf) $\tau_s(k+1) \rightarrow \tau_{s'}(k')$ whenever $\tau_s(k)=\tau_{s'}(k')$ and $k < k'$ (equivalently $k+1\leq k'$), indicating that shelf $s$ must be at $\tau_s(k+1)$ no later than $s'$ reaches $\tau_{s'}(k')$. For 1-robust $\mathcal{T}$, step indices strictly increase along Type-2 arcs ($k+1 < k'$). Fig.\ref{fig:CREST} (a) shows an example. \label{para:build-dg}

\noindent\textbf{Property~1}
\label{lem:tpg-acyclic}
\textit{The dependency graph $D$ constructed from any 1-robust MAPF shelf plan $\mathcal{T}$ is acyclic.}

\noindent\textit{Reason.}
By construction, the step index strictly increases along every arc for 1-robust $\mathcal{T}$.\qed

\noindent\textbf{Property~2}
\label{lem:tpg-gaurantee}
\textit{For any acyclic $D$ constructed from a MAPF shelf plan $\mathcal{T}$, if all events are executed in an order that respects every precedence arc, then the resulting execution is collision-free.}

\noindent\textit{Reason.}
Let $t(\tau_s(k))$ denote the execution time step of event $\tau_s(k)$. A Type-2 precedence $\tau_s(k+1) \rightarrow \tau_{s'}(k')$ prevents vertex collisions at $v=\tau_s(k)=\tau_{s'}(k')\in V$ since $t(\tau_s(k))<t(\tau_s(k+1))\leq t(\tau_{s'}(k'))$. Assume an edge collision occurs along $(u,v)=(\tau_s(k), \tau_{s}(k+1))=(\tau_{s'}(k'+1), \tau_{s'}(k'))\in E$, implying $t(\tau_s(k))+1=t(\tau_{s'}(k'))+1=t(\tau_s(k+1))=t(\tau_{s'}(k'+1))$. This contradicts the Type-2 precedence---either (if $k+1<k'$) $\tau_s(k+2)\rightarrow \tau_{s'}(k')$, indicating $t(\tau_s(k+1))<t(\tau_{s'}(k'))$, or (if $k'+1<k$) $\tau_{s'}(k'+2)\rightarrow \tau_{s}(k)$, indicating $t(\tau_{s'}(k'+1))<t(\tau_{s}(k))$. \qed

\subsection{Well-Formedness}
Following~\cite{li2023double}, we focus on \textit{well-formed} DD-MAPD instances in which $G$ remains connected after the removal of the initial locations of any $N-1$ agents, and a safe 1-robust MAPF shelf plan exists. These two conditions are sufficient to guarantee solvability and are not overly restrictive in practice \cite{li2023double} since a safe 1-robust shelf plan exists whenever at least two vertices of $G$ are initially empty \cite{luna2011push} (i.e., not the initial location of any agent and not occupied by any shelf). Such spatial slack is typically available in warehouse layouts.

\section{The CREST Framework}

CREST retains the two-stage decomposition of \cite{li2023double} but substantially improves the shelf execution by exploiting structural properties of the dependency graph. 
Executing shelf trajectories is inherently offline: full system information, such as when agents finish current tasks and when shelves become unconstrained, is available and can be leveraged to improve coordination. Moreover, shelves follow only the spatial components of their planned space-time trajectories, as limited agents can carry only a subset of shelves at once and the actual execution often deviates from the original schedule. In \cite{li2023double}, agents are assigned and planned only when dependencies are released, often leaving many agents idle while waiting for constraints to clear. CREST addresses this inefficiency by proactively releasing precedence constraints and coordinating agent-shelf interactions using full system information. The framework maintains the scalability of the decomposition while reducing idle time and interruptions. It further integrates three complementary strategies for constraint release and trajectory refinement to improve overall execution quality.

\subsection{Algorithm and Pseudocode}

\begin{figure}[t!]
    \centering
    \includegraphics[width=\columnwidth]{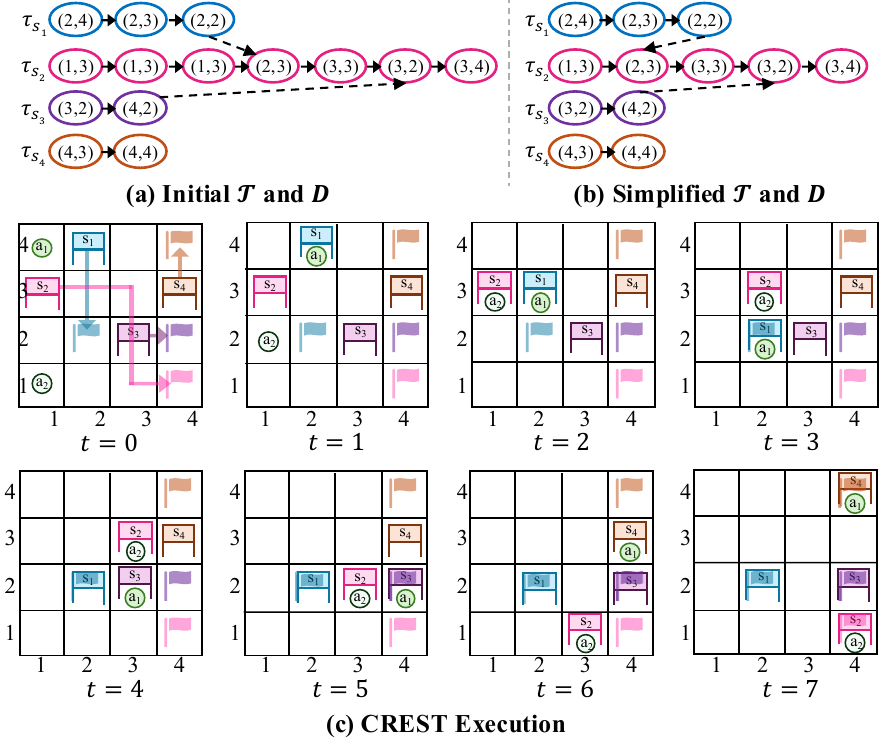}
    \caption{\jiaqi{\textbf{CREST running example with two agents ($a_1$, $a_2$) and four shelves ($s_1$–$s_4$)}. (a) shows the initial $\mathcal{T}$ and $D$ at $t=0$. Each shelf trajectory in $\mathcal{T}$ is an ordered waypoint sequence, and $D$ is a precedence graph defined over these waypoints; solid arrows denote Type-1 arcs and dashed arrows denote Type-2 arcs. (b) shows the simplified $\mathcal{T}$ and $D$ at $t = 0$, obtained by merging consecutive waypoints for each shelf. (c) shows the final execution produced by \framework~from $t = 0$ to $t = 7$.}}
    \label{fig:CREST}
     \vspace{-15pt}
\end{figure}

Algorithm~\ref{crest} outlines CREST. The blue lines indicate optional constraint-release strategies.
\jiaqi{CREST begins by calling $\textsc{BuildDep()}$ to construct $D$ and simplify $\mathcal{T}$. This function first initializes the dependency graph $D$ by adding Type-1 edges along each shelf trajectory waypoint in $\tau_s$ and Type-2 edges based on the inter-shelf location traversal precedence induced by $\mathcal{T}$, (Fig.~\ref{fig:CREST}(a)), detailed arc construction rules are described earlier in Sec.~\ref{para:build-dg}.
It then simplifies $\mathcal{T}$ and $D$ by merging consecutive waypoints at the same location (Fig.~\ref{fig:CREST}~(b)) [Line~\ref{crest:init}]. All agents are initially inactive [Line~\ref{crest:init_unlocked}].
}


\noindent\textbf{System Information:}
CREST tracks the following variables during execution:

\begin{itemize}
    \item \underline{$a.\text{current}$}: the \textbf{current location} of agent $a$, i.e., the last location of its path $\pi_a$.
    \item \underline{$t^{\text{avail}}_a$}: \textbf{available time} of agent $a$, defined as the end time of $\pi_a$.
    \item \underline{$s.\text{current}$}: the \textbf{current waypoint} of shelf $s$.
    \item \underline{$s.\text{next}$}: the \textbf{next waypoint} of shelf $s$.
    \item \underline{$t^{\text{rel}}_{s.\text{next}}$}: \textbf{shelf next-waypoint release time}, defined as the earliest time at which any agent can start carrying shelf $s$ from $s.\text{current}$ to $s.\text{next}$.
(1) If $s.\text{next}$ is \textit{constrained} (i.e., has incoming Type-2 arcs in $D$), $t^{\text{rel}}_{s.\text{next}}$ becomes finite only after all predecessor shelves $s'$ at the same location have passed $s.\text{next}$, and is defined as the maximum of the latest such passing time and the end time of $\phi_s$; otherwise, $t^{\text{rel}}_{s.\text{next}}=\infty$.
(2) If $s.\text{next}$ is not constrained, $t^{\text{rel}}_{s.\text{next}}$ equals the end time of $\phi_s$.
    \item \underline{$\hat{t}^{\text{start}}_{a,s}$}: the \textbf{estimated earliest start time} for $a$ to carry $s$ toward $s.\text{next}$, 
    $\hat{t}^{\text{start}}_{a,s} = \max\!\left(t^\text{avail}_a+dist(a.\text{current},s.\text{current}), t^\text{rel}_{s.\text{next}}\right)$

\end{itemize}

To begin with, whenever uncompleted shelves exist [Lines~\ref{crest:exist_shelf}-\ref{crest:update_D}], CREST calls $\textsc{ShelfAssignment()}$ to assign one agent-shelf pair. The function considers each candidate shelf $s$ with a finite release time $t^\text{rel}_{s.\text{next}}$ and computes a minimum-cost matching between agents and candidate shelves via the Hungarian method. 
The cost of matching agent $a$ to shelf $s$ equals the estimated waiting delay $\hat{t}^\text{start}_{a, s}-t^\text{rel}_{s.\text{next}} = \max\!\left(t^\text{avail}_a+dist(a.\text{current},s.\text{current}) - t^\text{rel}_{s.\text{next}}, 0\right)$, where $\textit{dist}(\cdot,\cdot)$ is the shortest-path distance on graph $G$.

\jiaqi{Unmatched agents or shelves incur a large penalty to prevent the Hungarian algorithm from favoring unmatched assignments for cost reduction.} Among the matched pairs, the function returns the pair $(a^*,s^*)$ with the smallest estimate delay $\hat{t}^\text{start}_{a, s}$ to maximize the potential for releasing future constraints and marks $a^*$ as \textit{active}, i.e., with the potential for carrying $s^*$ further [Line~\ref{crest:assign}].

CREST then iteratively processes all active agents whose shelves are uncompleted [Lines~\ref{crest:exist_unlock}–\ref{crest:unlock_false}], always selecting the agent $a$ with the earliest available time. If $a$ is newly assigned or $a$'s assigned shelf $s$ is available for execution no later than $a$'s available time, CREST calls MLSIPP (Multi-Label Safe Interval Path Planning)~\cite{tang2025large} to plan a path for $a$ to first move from its current location to $s.\text{current}$ and then carry $s$ along its unconstrained trajectory segment $\tau_s$ up to a \textbf{new waypoint} $s.\text{new}$ preceding the next constrained waypoint [Line~\ref{crest:plan_path}]. The resulting path is used to update $\pi_a$ and $\phi_s$, and all visited waypoints determine the exact times when dependent shelves become executable, allowing release times of relevant constrained waypoints to be updated accordingly. All agents with assigned shelves are marked \textit{active} for consideration in future iterations [Line~\ref{crest:unlock_true}]; however, without \textsc{DepSwitch} and \textsc{GroupReplan} activated, this consideration is effective only for agents whose shelves have newly released constraints, as others will be inactive again when checked later [Line~\ref{crest:unlock_false}]. When no further activation is possible, CREST identifies the earliest end time among all shelf paths in $\Phi$ and removes all Type-2 arcs in $D$ that originate from waypoints traversed no later than this time. This is safe because all subsequent waypoints—whether already planned in $\Phi$ or to be planned in future iterations— satisfy the removed precedence constraints, thereby improving efficiency [Line~\ref{crest:update_D}].

\begin{algorithm}[t]
\caption{CREST}\label{crest}
\KwIn{initial safe 1-robust MAPF shelf plan $\mathcal{T}$} \label{crest:input}
\KwOut{agent paths $\Pi$, shelf paths $\Phi$} \label{crest:output}
$D \gets \textsc{BuildDep}(\mathcal{T})$, $\Pi \gets \{\langle \pi_a(0) \rangle\}_a$, $\Phi \gets \{\langle p_s \rangle\}_s$\; \label{crest:init}
$\forall a: a.\text{active} \gets \text{False}$\; \label{crest:init_unlocked}
\While{uncompleted shelves exist}{ \label{crest:exist_shelf}
    call $\textsc{ShelfAssignment()}$ to assign $(a^*,s^*)$ and $a^*.\text{active} \gets \text{True}$\; \label{crest:assign}
    \While{$a \gets \argmin\limits_{a: a.\text{active} \wedge a\text{'s shelf uncompleted}} t^\text{avail}_{a'}$ \textbf{exists}}{ \label{crest:exist_unlock}
        \If{$a=a^*$ \textbf{or} $a$'s shelf $s$: $t^\text{rel}_{s.\text{next}}\leq t^\text{avail}_a$ \color{blue}
        \textbf{or} $t^\text{rel}_{s.\text{next}} = \infty \land \bigl[\textsc{DepSwitch}(s.\text{next}, D, \Pi, \Phi) 
        \lor \textsc{GroupReplan}(s, D, \Pi)\bigr]$}{ \label{algo-CoMAWR:line11}
            {\color{blue}\If{\textsc{SingleReplan}}{\label{crest:str}
            $\textsc{ExtractMAPF}(\mathcal{T}, D, \Pi, \Phi)$\;\label{crest:str:extract}
            $\textsc{MinCostMaxCarry}(s, \mathcal{T}, D)$\; \label{crest:str:plan}
            }}
            call $\textsc{MLSIPP}()$ to plan a path for $a$ to carry $s$ and update $\Pi$ and $\Phi$ accordingly\;\label{crest:plan_path} 
            $\forall$ assigned $a': a'.\text{active} \gets \text{True}$\; \label{crest:unlock_true}
        }
        \Else{ \label{algo-CoMAWR:line15}
            $a.\text{active} \gets \text{False}$\; \label{crest:unlock_false}
        }
    update $D$ based on $\Phi$\; \label{crest:update_D}
    }
}

\end{algorithm}

\noindent\textbf{Path Planning Details:} MLSIPP is an A*-based planner that computes a time-minimal, collision-free path through a sequence of goals. Each state encodes a location, a maximal safe time interval (collision-free with other agents’ paths), and a progress label along the goal sequence.
\jiaqi{Compared to vertex–timestep–based planning (e.g., MLA*), MLSIPP searches over vertex–interval states, which typically leads to a smaller search space and improved efficiency in practice \cite{phillips2011sipp,li2022mapf}.} CREST employs MLSIPP when planning a path for agent $a$ to carry shelf $s$, with two key modifications:
\begin{enumerate}[nosep,leftmargin=*]
    \item The initial location $\pi_a(0)$ is appended to the end of the goal sequence (after $s.\text{new}$) to form a ``dummy path'' segment from the new waypoint $s.\text{new}$ back to $\pi_a(0)$. This ensures completeness for all well-formed instances. The ``dummy path'' is treated external (not stored in $\pi_a$) but must be avoided by other agents during their planning.
    \item During the second label (between $s.\text{current}$ and $s.\text{new}$), state expansions include additional wait and move-backward actions along the unconstrained trajectory segment, whereas \cite{li2023double} permits only move-forward actions. These additional actions still satisfy all precedence constraints in $D$, ensuring collision-free shelf execution (Property 2) while providing greater flexibility in local coordination.
\end{enumerate}
Agents and shelves commit to following their planned paths $\Pi$ and $\Phi$, allowing CREST to precisely track agent available times and waypoint execution times---information crucial for subsequent shelf assignment [Line~\ref{crest:assign}] and path planning [Line~\ref{crest:plan_path}].

\jiaqi{
Fig.~\ref{fig:CREST} shows a running example of CREST in the no-overhead setting, without additional constraint-release strategies. 
From the initial $\mathcal{T}$, a simplified $\mathcal{T}$ and $D$ are extracted. 
In Round~1 of Algo.~\ref{crest} [Lines~3–14], shelf $s_1$ is assigned to agent $a_1$, which carries it to the goal via $\pi_{a_1}(0)\sim\pi_{a_1}(5) = [(1,4),(2,4),(2,3),(2,2),(3,2)]$, releases the waypoint of $s_2$ at $(2,3)$ at timestep~2. In Round~2, shelf $s_2$ is assigned to agent $a_2$. Since the waypoint of $s_2$ at $(3,2)$ is constrained by $s_3$, agent $a_2$ plans a path only up to waypoint $(3,3)$ and then waits, with $\pi_{a_2}(0)\sim\pi_{a_2}(6) = [(1,1),(1,2),(1,3),(2,3),(3,3),(4,3)]$. In Round~3, shelf $s_3$ is assigned to agent $a_1$, which reaches its goal and activates $a_2$ [Line~11]. Agent $a_2$ then completes the delivery of $s_2$. In the last round, shelf $s_4$ is assigned to agent $a_1$ and delivered. Without such coordination by CREST, agent $a_2$ would initially move toward $s_3$ and later switch to $s_4$ when $s_2$ becomes blocked, resulting in inefficient execution.
}

\noindent\textbf{Theorem~1 (Completeness)}
\label{thm:completeness}
\textit{CREST solves all well-formed DD-MAPD instances.}

\noindent\textit{Proof.}
The dependency graph $D$ constructed from the safe 1-robust $\mathcal T$ is acyclic by Property~1. CREST only removes Type-2 arcs made irrelevant by already-traversed waypoints, thus preserving acyclicity. At any time, let $D'$ denote the subgraph of $D$ induced by untraversed waypoints. Since $D'$ is acyclic, it has at least one source waypoint, which must be the first untraversed waypoint of some shelf $s$, i.e., $s.\text{next}$. This waypoint is either unconstrained or has incoming Type-2 arcs in $D$ only from traversed waypoints (otherwise, it would not be a source in $D'$), and thus has a finite release time. Consequently, $\textsc{ShelfAssignment()}$ always finds a feasible pair $(a,s)$. On Line~\ref{crest:plan_path}, MLSIPP guarantees to find a collision-free path for $a$ to carry $s$ from $s.\text{current}$ to $s.\text{new}$ since such a path always exists: $a$ can first follow its dummy path segment and wait at its initial location until all other agents complete their paths (including dummy path segments) and remain at their initial locations indefinitely. By well-formedness and the safe shelf plan, $a$ can then move to $s.\text{current}$, carry $s$ along $\tau_s$ up to $s.\text{new}$, and finally return to its initial location along a new dummy path segment. Moreover, since every planned shelf path segment respects all precedence constraints, Property 2 ensures that all shelf paths are collision-free. Therefore, in each iteration of the outer loop [Lines~\ref{crest:exist_shelf}-\ref{crest:update_D}], at least one additional unconstrained trajectory segment is assigned and executed, and all shelves eventually reach their delivery locations. \qed

\noindent\textbf{Handling Lift/Place Overhead:} Incorporating a fixed lift/place overhead $\Delta$ requires only minor modifications to Algorithm~\ref{crest}. On Line~\ref{crest:exist_unlock}, the condition for deciding whether an agent $a$ should continue carrying its shelf $s$ is updated to $t^\text{rel}_{s.\text{next}}\leq t^\text{avail}_a + 2\Delta$, which accounts for the minimum additional cost of switching $a$ to another shelf when $s$ cannot be advanced immediately. MLSIPP can naturally incorporate $\Delta$ whenever $a$ lifts or places $s$. Likewise, calculations of available times and assignment costs can directly incorporate this overhead, depending on whether agents continue carrying or switch shelves, without affecting completeness or collision guarantees.

\begin{figure}[t!]
\centering
\includegraphics[width=0.96\columnwidth]{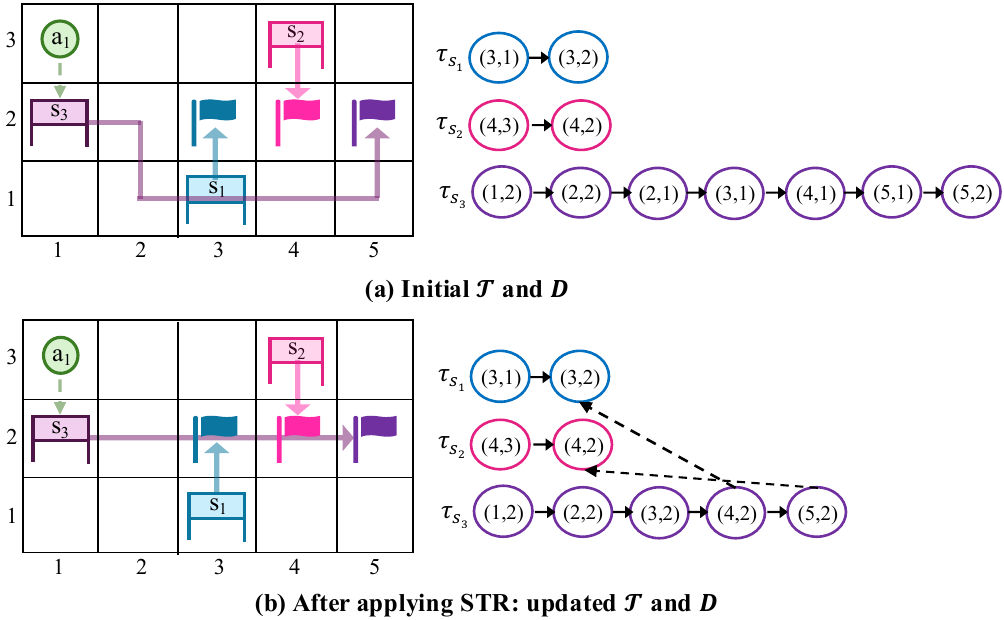}
\caption{\jiaqi{STR example with one agent ($a_1$) and three shelves ($s_1$, $s_2$, $s_3$). With $a_1$ assigned to $s_3$, the baseline detours to avoid $s_1$ and $s_2$ (left). STR replans the remaining path of $s_3$ to shorten the unexecuted segment and avoid being blocked at $(3,1)$ (right).}}
\label{fig:STR}
\vspace{-10pt}
\end{figure}

\subsection{Single Trajectory Replanning (STR)}

\jiaqi{As illustrated in Fig.~\ref{fig:STR}, STR leverages partial execution to shorten and unconstrain the remaining path of $s_3$, enabling it to move directly toward its delivery location. STR implements this execution-aware refinement in two steps. }
First, it calls $\textsc{ExtractMAPF}()$ to reconstruct unexecuted trajectories as a time-aligned MAPF shelf plan, updating $\mathcal{T}$. In this reconstruction: (1) all shelves follow their current paths $\Phi$; (2) any unassigned shelf whose path ends earlier than the earliest agent available time $t^\text{avail}_{a^*}$ waits at its last location until $t^\text{avail}_{a^*}$ (since it cannot be carried sooner); and (3) thereafter, each shelf proceeds along its untraversed waypoints, advancing when unconstrained and waiting otherwise, thus respecting all dependency arcs in $D$. 
STR then calls $\textsc{MinCostMaxCarry}()$ to run an A$^*$ search over location-step pairs to replan $\tau_s$ from its last traversed waypoint to the delivery location $d_s$ so that it is collision-free with other trajectories and safe (not using agent initial locations), while minimizing its arrival step at $d_s$ with tie-breaking on maximizing carrying duration---the number of steps up to the first location already visited earlier by another trajectory. If the resulting dependency graph remains acyclic, STR updates $\tau_s\in \mathcal{T}$ and $D$; otherwise, it restores the prior $\mathcal{T}$ and $D$, preserving the completeness of CREST.


\subsection{Dependency Switching (DS)}

\begin{figure}[t!]
\centering
\includegraphics[width=\columnwidth]{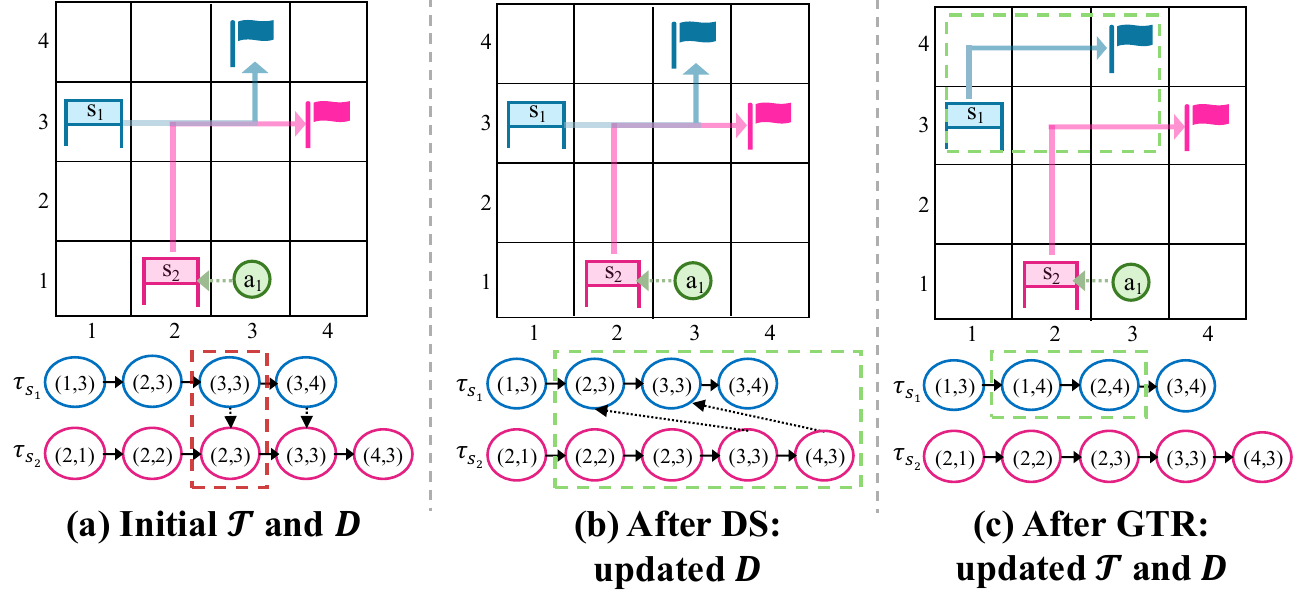}
\caption{\jiaqi{DS and GTR example with one agent ($a_1$) and two shelves ($s_1$ and $s_2$). Agent $a_1$ carries $s_2$ first. When $s_2$ is constrained by $s_1$ at location $(2,3)$, DS reverses the corresponding Type-2 dependencies to allow $s_2$ to proceed first, whereas GTR replans the unexecuted portion of $s_1$’s path to free the route for $s_2$.}}
\label{fig:SW-GTR}
\vspace{-5pt}
\end{figure}

During agent activation (Line~\ref{crest:exist_unlock} of Algorithm~\ref{crest}), DS attempts to reverse precedence dependencies when agent $a$’s carried shelf $s$ is constrained at $s.\text{next}=\tau_s(k)$ with an infinite release time so that $a$ can continue carrying $s$ instead of switching shelves. DS applies a recursive arc reversal procedure that replaces each Type-2 incoming $\tau_{s'}(k')\rightarrow\tau_s(k)$ in $D$ with $\tau_s(k+1)\rightarrow\tau_{s'}(k'-1)$, effectively prioritizing $s$ over $s'$. The procedure terminates unsuccessfully if $\tau_{s'}(k'-1)$ has already been traversed. 
Since such reversals can introduce cycles, DS explores alternative reversals via a depth-first recursive search, branching over all Type-2 arcs in any detected cycle, and prunes branches at a predefined depth limit (empirically set to 5 in our setting) to keep computation lightweight, as successful dependency switches rarely require more than a few reversals in practice. 
If any recursive branch yields an acyclic dependency graph with an estimated makespan no larger than $D$, the reversal branch is accepted, and $D$ is updated; otherwise, the branch is pruned. The makespan of any dependency graph $D'$ can be estimated by (1) setting earlier-ending unassigned shelves to start at the earliest agent available time $t^\text{avail}_{a^*}$ and others to start at their current path end times in $\Phi$; and (2) propagating earliest feasible execution times through $D'$ assuming shelves move by themselves while respecting precedence constraints. DS preserves the acyclicity of $D$ and thus the completeness of CREST, while advancing constrained shelves whenever beneficial. \jiaqi{Fig.~\ref{fig:SW-GTR}(a) and (b) present a simple situation where applying DS enables continued carrying of a shelf, resulting in improved execution efficiency.}

\subsection{Group Trajectory Replanning (GTR)}
\jiaqi{GTR can be applied either independently or as a fallback when DS fails to reverse the dependency
$\tau_{s'}(k') \rightarrow \tau_s(k)$.
In both cases, GTR replans the unexecuted segments of all constraining $\tau_{s'}$
so that none uses $s.\text{next} = \tau_s(k)$ earlier than $s$.}
GTR first verifies that all constraining shelves $s'$ are unassigned; if any is assigned or has already traversed a waypoint constraining $s.\text{next}$, it terminates unsuccessfully. It then calls $\textsc{ExtractMAPF}()$ to reconstruct unexecuted trajectories as a time-aligned MAPF shelf plan (as in STR), updating $\mathcal{T}$, and identifies the constrained $s.\text{next}$ with its step $k$ and all constraining shelves $s'$ in the reconstruction. Each $\tau_{s'}$ is replanned using an A$^*$ search over location-interval pairs (a single-label MLSIPP) from its last traversed waypoint to the delivery location $d_{s'}$ so that it avoids $s.\text{next}$ before step $k$ and is collision-free with other trajectories and safe (not using agent initial locations), while minimizing its arrival step at $d_{s'}$. If all replannings succeed, none increases any arrival step, the resulting dependency graph remains acyclic, and its estimated makespan (computed as in DS) does not exceed that of $D$, then GTR updates $\mathcal{T}$ and $D$. Otherwise, it restores the prior $\mathcal{T}$ and $D$, preserving the completeness of CREST.  \jiaqi{Fig.~\ref{fig:SW-GTR} (a) and (c) shows the resulting $\mathcal{T}$ and $D$ after applying GTR.}

\section{Empirical Evaluation}

We conduct all experiments on a server with a 16-core CPU and 8 GB RAM. Both CREST and the MAPF-DECOMP baseline~\cite{li2023double} are implemented in C++.
\jiaqi{We do not compare against makespan-optimal solvers, as their target domain differs substantially from ours (e.g.,~\cite{sherma2025agent} considers at most 16 shelves, and we could not reproduce its results on 16-shelf instances within 30 minutes), whereas our evaluation focuses on large-scale warehouse scenarios with hundreds to thousands of shelves. Instead, we choose MAPF-DECOMP(PP) as the baseline, since~\cite{li2023double} identifies the PP variant as the strongest, achieving 2-6$\times$ faster runtime than other variants with at most 1\% degradation in makespan and flowtime.}
As the original MAPF-DECOMP implementation is unavailable, we re-implement the PP variant—complete for well-formed instances—and replace its single-agent planner with SIPP for efficiency. All methods use the same precomputed shelf plan and a 10,000 s runtime limit per instance.

\begin{figure}[t!]
\centering
\includegraphics[width=\columnwidth]{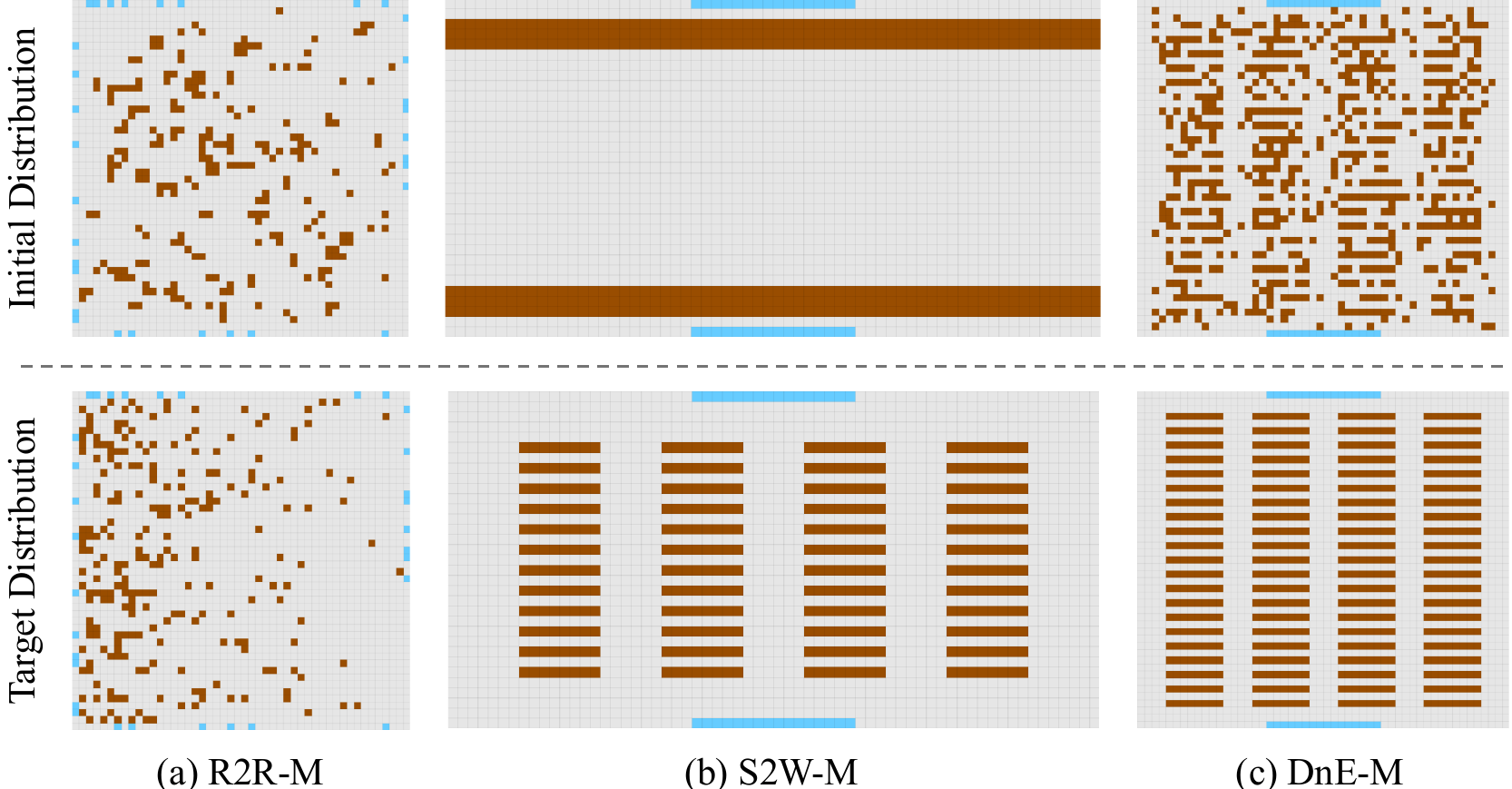}
\caption{Distributions of shelf pickup (top) and delivery (bottom) cells (brown) and agent initial cells (blue) across three layouts.}
\label{fig:warehouse-layout}
\vspace{-15pt}
\end{figure}

\begin{table}[t!]
\caption{Layout Specifications.}
\vspace{-5pt}
\label{table-instance-summary}
\centering
\Huge\scalebox{0.45}{%
\begin{tabular}{c|c|c|c}
\toprule
\textbf{Map} & \textbf{Size} & \textbf{\# Shelves ($\frac{\text{\# Shelves}}{\text{\# Cells}}$)} & \begin{tabular}{@{}c@{}} \textbf{\# Rearranged Shelves}\\\textbf{($\frac{\text{\# Rearranged Shelves}}{\text{\# Cells}}$)} \end{tabular} \\
\midrule
\rTwoR-M   & 48$\times$48   & 460 (20.0\%)   & 230 (10.0\%) \\
\sTwoW-M   & 64$\times$33   & 384 (18.2\%)   & 384 (18.2\%) \\
\dAndE-M   & 47$\times$45   & 672 (31.7\%)   & 403 (19.0\%) \\
\rTwoR-L   & 96$\times$96   & 1,843 (20.0\%)  & 921 (10.0\%) \\
\sTwoW-L   & 128$\times$111 & 2,816 (19.8\%)  & 2,816 (19.8\%) \\
\dAndE-L   & 128$\times$111 & 2,816 (19.8\%)  & 2,816 (19.8\%) \\
\bottomrule
\end{tabular}
}
\vspace{-5pt}
\end{table}

\subsection{Environment Setup}
\noindent\textbf{Shelf Layouts:} 
We evaluate three representative layout types (Fig.\ref{fig:warehouse-layout}):
(1) Random-to-Random (\textbf{\rTwoR}): adopted from \cite{li2023double}, where shelves are randomly distributed and rearranged to another random configuration;
(2) Staging-to-Warehouse (\textbf{\sTwoW}): modeling practical scenarios where incoming shelves temporarily stored in a staging zone are relocated to designated storage locations, \jiaqi{with each shelf potentially ending up in any of the designated locations.}
(3) Distributed-and-Exchange (\textbf{\dAndE}): modeling in-operation rearrangement that relocates all shelves to designated storage locations, with half initially placed randomly within these storage zones and the rest at random non-storage locations.
We construct a medium (\textbf{-M}) and a large (\textbf{-L}) layout for each type and generate 25 instances per layout. Table.\ref{table-instance-summary} summarizes the map sizes, shelf counts, and numbers of rearranged shelves (those with distinct pickup and delivery locations).  

\noindent\textbf{Trajectory Generation:}  
To guarantee completeness, we adapt existing MAPF solvers to generate safe 1-robust shelf plans for all instances. For \sTwoW~and \dAndE, we adapt the anytime MAPF-LNS2~\cite{li2022mapf} with a 600s runtime limit. For other layouts, we adapt EECBS~\cite{li2021eecbs} with a suboptimality factor of 1.6, following \cite{li2023double}. 


\noindent\textbf{Lift/Place Overhead:} We evaluate both the original DD-MAPD setting without lift/place overhead \cite{li2023double} and an extended setting with a unit-time cost ($\Delta$=1) for each lift or place action to better capture real operation times.

\subsection{Evaluation Metrics}
We evaluate solution quality using the sum of costs and makespan, reported in their normalized form to better capture the execution effectiveness independent of instance scale and the initial shelf plan $\mathcal{T}$. Normalization removes the baseline contribution of $\mathcal{T}$, highlighting improvements in execution coordination. The normalized sum of cost (\textbf{Norm. Cost}) is $\sum_{i=1}^{N} |\pi_{a_i}| - \sum_{j=1}^{M}|\tau_{j_i}|$. The normalized makespan (\textbf{Norm. Mksp}) is $\max_{i=1,\ldots,N} |\pi_{a_i}| - {\sum_{j=1}^{M} |\tau_{j_i}|}/{N}$, where the second term represents the idealized lower bound on makespan assuming all shelves move on their own but are limited by the number of agents. The average number of lift-place transitions per shelf (\textbf{\# Switch/Shelf}) reflects how frequently a shelf changes its carrying agent, quantifying the benefit of constraint release and continuous carrying. The reported runtime excludes the time for generating the initial shelf plan, which is shared by all methods.

\subsection{Results} 
\jiaqi{This section analyzes results across different frameworks and overhead settings; complete numerical results, including additional unnormalized metrics for both with- and without-overhead settings, are provided in the technical report.}



\begin{table}[t]
\caption{Results for the MAPF-DECOMP baseline and CREST without overhead. Percentages indicate reductions relative to the baseline. `$N$' denotes the number of agents.}
\label{table-crest-baseline}
\renewcommand{\arraystretch}{0.9}
\centering
\Huge\scalebox{0.35}{%
\begin{tabular}{c|c|c|c|c|c}
\toprule
\textbf{Map} & \textbf{Method} 
& \textbf{Norm. Cost}
& \textbf{Norm. Mksp}
& \textbf{\# Switch/Shelf}
& \textbf{Time (s)} \\
\midrule
\multirow{2}{*}{\shortstack{\rTwoR-M \\ $N$=32}}
 & Baseline & 7,575.40 & 316.75 & 1.48 & 9.47 \\
 & CREST & 5,863.36 (-22.6\%) & 287.51 (-9.2\%) & 1.27 (-14.2\%) & 6.88 \\
\midrule
\multirow{2}{*}{\shortstack{\sTwoW-M \\ $N$=32}}
 & Baseline & 17,662.92 & 648.46 & 5.37 & 21.74 \\
 & CREST & 13,920.84 (-21.2\%) & 515.11 (-20.6\%) & 5.00 (-6.8\%) & 15.75 \\
\midrule
\multirow{2}{*}{\shortstack{\dAndE-M \\ $N$=32}}
 & Baseline & 14,504.16 & 547.56 & 3.13 & 24.64 \\
 & CREST & 11,820.40 (-18.5\%) & 465.76 (-14.9\%) & 2.84 (-9.1\%) & 22.32 \\
\midrule
\multirow{2}{*}{\shortstack{\rTwoR-L \\ $N$=100}}
 & Baseline & 72,018.48 & 915.57 & 3.48 & 341.43 \\
 & CREST & 54,331.84 (-24.6\%) & 798.77 (-12.8\%) & 3.09 (-11.3\%) & 347.89 \\
\midrule
\multirow{2}{*}{\shortstack{\sTwoW-L \\ $N$=100}}
 & Baseline & 461,249.72 & 4,760.87 & 16.29 & 6,258.26 \\
 & CREST & 347,397.56 (-24.7\%) & 3,640.87 (-23.5\%) & 15.27 (-6.3\%) & 8,352.40 \\
\midrule
\multirow{2}{*}{\shortstack{\dAndE-L \\ $N$=100}}
 & Baseline & 334,300.36 & 3,466.45 & 18.01 & 4,899.67 \\
 & CREST & 280,740.56 (-16.0\%) & 2,949.21 (-14.9\%) & 16.69 (-7.3\%) & 8,634.67 \\
\bottomrule
\end{tabular}}
 \vspace{-10pt}
\end{table}

\noindent\textbf{CREST vs. Baseline:} 
As shown in Table~\ref{table-crest-baseline}, \framework~consistently outperforms the MAPF-DECOMP baseline across all settings. It reduces normalized cost by 16--24.7\%, normalized makespan by 9.2--23.5\%, and number of lift/place actions by 6.3--14.2\%.
These improvements confirm that minimizing inter-shelf agent travel and reducing unnecessary shelf switching substantially enhances execution effectiveness.
\framework's runtime increases moderately on large maps because MLSIPP considers additional wait and move-backward actions along unconstrained segments, which enlarge the search space. This overhead is acceptable in the context of offline planning, where improving overall system throughput by reducing agent idle time and enabling continuous shelf carrying outweighs marginal increases in computation.

\begin{figure}[t]
    \centering  
    \includegraphics[width=\columnwidth]{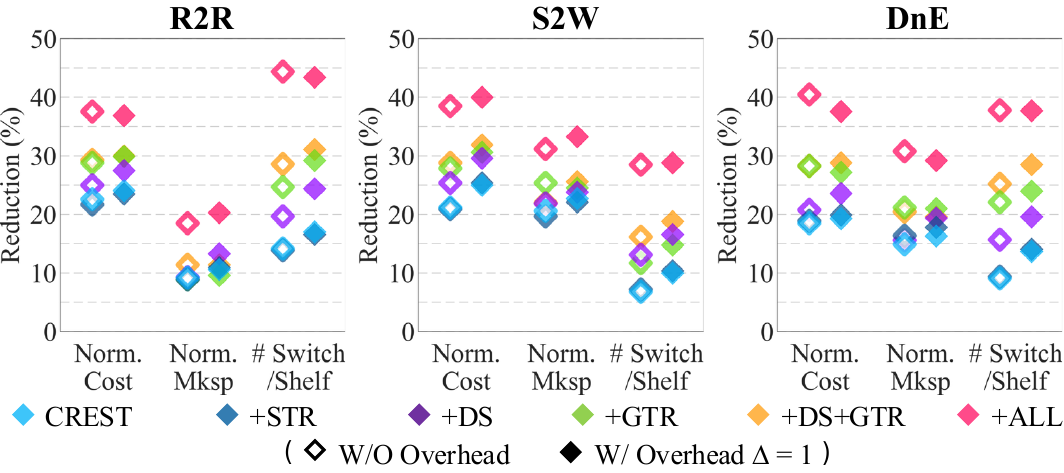}
    \caption{\jiaqi{Percentage reduction over baseline achieved by CREST and its variants on three layouts, measured by normalized sum of costs, normalized makespan, and number of lift/place transitions per shelf.}}
    \label{fig:all-result}
 \vspace{-5pt}
\end{figure}

\noindent\textbf{Ablation on Constraint-Release Strategies:}
As shown in Fig.~\ref{fig:all-result}, effectiveness of each strategy varies when applied individually within \framework. 
\trajReplan\ is the most effective overall when used alone, reducing the normalized cost, normalized makespan, and shelf switching by up to 30.6\%, 25.4\%, and 29.2\%, respectively. 
\switch\ ranks second, offering smaller reductions than \trajReplan~but with considerably less runtime \jiaqi{(see technical report for runtime details).}
\singleAgentTrajReplan ~yields only marginal gains when used alone, yet when combined with the other two strategies (+All), it contributes significantly, further improving upon \trajReplan+\switch~by up to 12.2\%, 10.7\%, and 12.5\% in the three metrics. 
Overall, applying all three strategies achieves the best performance across all scenarios, improving over the baseline by up to 40.5\%, 33.3\%, and 44.4\% in the three metrics. 
These results demonstrate that execution-aware constraint release, which promotes continuous shelf carrying, effectively lowers path costs and shortens overall rearrangement completion time.

\begin{figure}[t!]
\centering
\includegraphics[width=\columnwidth]{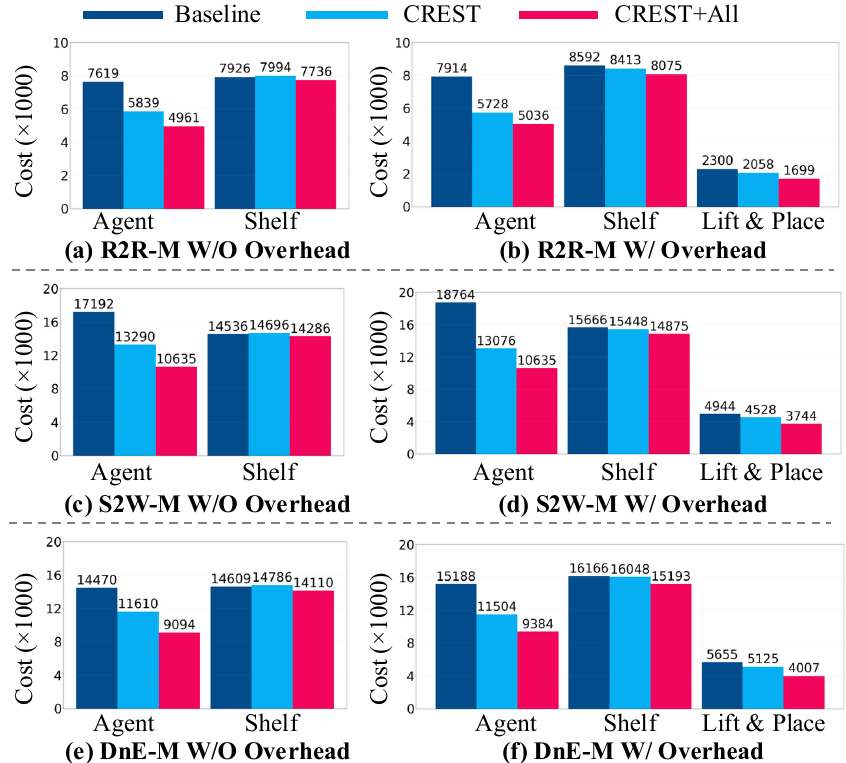}
\caption{Cost breakdown for three methods on medium layouts.}
\label{fig:DnE-breakdown}
\end{figure}

\noindent\textbf{Ablation on Lift/Place Overhead:} Introducing a unit-time for each lift/place action increases the normalized sum of costs and makespan, with minimal effect on shelf switching. With this overhead, \framework~and all constraint release strategies achieve larger absolute reductions in both cost and makespan (Fig.~\ref{fig:DnE-breakdown}), while maintaining consistent relative improvements over the baseline \jiaqi{(Fig.~\ref{fig:all-result}, see the technical report for detailed numerical results)}. These results indicate that our proposed framework and strategies are robust and equally effective under both settings.

\noindent\textbf{Cost Breakdown:}
As shown in Fig.~\ref{fig:DnE-breakdown}, the (unnormalized) sum of costs can be decomposed into \textbf{Agent} Travel (agent motion without a carried shelf), \textbf{Shelf} Transport (agent motion while carrying in a shelf), and lift/place overhead.
\framework\ and all constraint-release strategies reduce the Agent Travel the most, with moderate gains in lift/place overhead (limited by dependencies already embedded in the initial shelf plan). STR and GTR also reduce Shelf Transport since they replan shelf trajectories, though the pre-optimized initial shelf plan leaves less room for improvement compared to reducing Agent Travel. 
While CREST follows the same initial shelf plan spatially as the baseline, it slightly increases Shelf Transport by adding wait and move-backward actions along unconstrained trajectory segments that are introduced to improve local coordination. Nevertheless, this minor overhead is outweighed by significant reductions in Agent Travel and lift/place overhead, resulting in a clear overall cost improvement.

\begin{table}[t!]
\caption{Runtime (s) per shelf, shelf step, and agent step.}
\vspace{-5pt}
\label{table:runtime-efficiency}
\centering
\Huge\scalebox{0.4}{%
\begin{tabular}{c|c|c|c|c|c|c}
\toprule
\multirow{2}{*}{\textbf{Map}} 
& \multicolumn{3}{c|}{\textbf{W/O Overhead}} 
& \multicolumn{3}{c}{\textbf{W/ Overhead ($\Delta$ = 1)}} \\
\cmidrule(lr){2-4}\cmidrule(lr){5-7}
& \multicolumn{1}{c|}{\textbf{/Shelf}} 
& \multicolumn{1}{c|}{\textbf{/Shelf Step}} 
& \multicolumn{1}{c|}{\textbf{/Agent Step}} 
& \multicolumn{1}{c|}{\textbf{/Shelf}} 
& \multicolumn{1}{c|}{\textbf{/Shelf Step}} 
& \multicolumn{1}{c}{\textbf{/Agent Step}}  \\
\midrule
\rTwoR-M   & 0.0299 & 0.0009 & 0.0005 & 0.0307 & 0.0009 & 0.0004 \\
\sTwoW-M   & 0.0410 & 0.0011 & 0.0006 & 0.0398 & 0.0011 & 0.0005 \\
\dAndE-M   & 0.0554 & 0.0015 & 0.0008 & 0.0556 & 0.0015 & 0.0007 \\
\rTwoR-L   & 0.3777 & 0.0055 & 0.0030 & 0.3538 & 0.0052 & 0.0024 \\
\sTwoW-L   & 2.9661 & 0.0294 & 0.0132 & 3.2029 & 0.0317 & 0.0124 \\
\dAndE-L   & 3.0663 & 0.0334 & 0.0160 & 3.3593 & 0.0366 & 0.0145 \\
\bottomrule
\end{tabular}}
\end{table}

\noindent\textbf{Online Applicability:}
As shown in Table \ref{table:runtime-efficiency}, planning for a single shelf is highly efficient---below 0.06s on medium layouts and 3.4s on large ones. Since CREST plans only for executing one trajectory segment at a time, each computed path is short. Estimating planning time from per-step computation (below 0.037s per shelf step and 0.016s per agent step) yields values even smaller than the measured per-shelf time, confirming that planning easily fits within real-time execution, where robots typically need over 1s per step. Although CREST currently operates offline, its low computational overhead makes it readily adaptable for real-time or online deployment in practical multi-robot warehouses.

\section{Conclusion}

We proposed CREST, a new execution framework for DD-MAPD that improves execution performance by proactive constraint release. By exploiting full system information, CREST reduces agent idle time and lift-place actions while maintaining the scalability and completeness of decomposed planning. Future work will focus on extending CREST to online settings with dynamic task arrivals, incorporating learning-based heuristics for dependency switching and trajectory replanning, and validating CREST on real multi-robot systems to demonstrate its effectiveness in practical deployments.

\bibliographystyle{IEEEtran}
\bibliography{reference}

\clearpage
\onecolumn
\section*{Technical Report}

\section{Additional Results}

In addition to the metrics reported in the main paper, we provide supplementary results using unnormalized metrics; all metrics are summarized below:

\begin{itemize}
    \item \textbf{Sum of Cost (Cost):} $\sum_{i=1}^{N} \lvert \pi_{a_i} \rvert$, representing the total agent travel distance required to rearrange all shelves, including both non-carrying travel distance (agents moving between shelves) and carrying travel distance.
    \item \textbf{Normalized Sum of Cost (Norm.\ Cost):} $\sum_{i=1}^{N} \lvert \pi_{a_i} \rvert - \sum_{j=1}^{M} \lvert \tau_j \rvert$, capturing non-carrying agent travel cost, including costs to pick up the next shelf and costs due to shelf switching.
    \item \textbf{Makespan (Mksp):} $\max_{i=1,\ldots,N} \lvert \pi_{a_i} \rvert$, denoting the total time required to complete the shelf rearrangement.
    \item \textbf{Normalized Makespan (Norm.\ Mksp):} $\max_{i=1,\ldots,N} \lvert \pi_{a_i} \rvert - \frac{1}{N}\sum_{j=1}^{M} \lvert \tau_j \rvert$, where the second term represents an idealized lower bound assuming shelves move independently but are constrained by the number of agents; this metric reflects makespan delay induced by agent travel.
    \item \textbf{\# Switch/Shelf:} the average number of lift--place transitions per shelf, measuring how frequently a shelf changes its carrying agent and quantifying the benefit of constraint release and continuous carrying.
\end{itemize}

\textit{Note: $\pi_{a_i}$ denotes the executed path of agent $a_i$, and $\tau_j$ denotes the planned trajectory of shelf $s_j$.}

Table.~\ref{table-main-result-no-overhead} and Table.~\ref{table-main-result-with-overhead} report detailed results for the settings without overhead and with unit overhead, respectively. We omit STDs for unnormalized metrics since they are identical to the STDs of the corresponding normalized metrics.
As shown in the tables, the reported standard deviations are moderate across instances, indicating that the observed performance trends are consistent. 
Across all settings, CREST consistently improves over the baseline while remaining efficient in practice, and the runtimes of both CREST and its strategy variants remain computationally tractable under our offline, large-scale evaluation setup.

\begin{table*}[b!]
\caption{Results for the baseline and all CREST variants \textbf{without overhead}. Values are reported as mean $\pm$ standard deviation over 25 instances for normalized metrics (Norm. Cost and Norm. Mksp); other metrics are reported as mean. Percentages denote reductions relative to the baseline. `$N$' denotes the number of agents in the setting.}
\vspace{5pt}
\label{table-main-result-no-overhead}
\centering
\begin{adjustbox}{max width=\textwidth}
\begin{tabular}{c|c|c|c|c|c|c|c}
\toprule
\multicolumn{1}{c|}{\textbf{Map + N}} &
\multicolumn{1}{c|}{\textbf{Method}} &
\multicolumn{1}{c|}{\textbf{Norm. Cost}} &
\multicolumn{1}{c|}{\textbf{Cost}} &
\multicolumn{1}{c|}{\textbf{Norm. Mksp}} &
\multicolumn{1}{c|}{\textbf{Mksp}} &
\multicolumn{1}{c|}{\textbf{\# Switch/Shelf}} &
\multicolumn{1}{c}{\textbf{Time (s)}} \\
\midrule
\multirow{7}{*}{\shortstack{\rTwoR-M \\ +32}}
 & Baseline   & 7,575.40 $\pm$ 726.55 & 15,545.00 & 316.75 $\pm$ 28.73 & 565.80 & 1.48 & 9.47 \\
 & \framework & 5,863.36 $\pm$ 512.80 (-22.6\%) & 13,832.96 (-11.0\%) & 287.51 $\pm$ 35.95 (-9.2\%) & 536.56 (-5.2\%) & 1.27 (-14.2\%) & 6.88 \\
 & $+$\singleAgentTrajReplan & 5,932.40 $\pm$ 629.03 (-21.7\%) & 13,902.00 (-10.6\%) & 288.43 $\pm$ 39.42 (-8.9\%) & 537.48 (-5.0\%) & 1.27 (-13.9\%) & 375.20 \\
 & $+$\switch & 5,683.00 $\pm$ 513.41 (-25.0\%) & 13,652.60 (-12.2\%) & 287.35 $\pm$ 37.47 (-9.3\%) & 536.40 (-5.2\%) & 1.19 (-19.7\%) & 184.53 \\
 & $+$\trajReplan & 5,394.96 $\pm$ 534.39 (-28.8\%) & 13,364.56 (-14.0\%) & 288.63 $\pm$ 35.47 (-8.9\%) & 537.68 (-5.0\%) & 1.11 (-24.7\%) & 445.55 \\
 & $+$\switch $+$\trajReplan & 5,356.60 $\pm$ 518.92 (-29.3\%) & 13,326.20 (-14.3\%) & 280.75 $\pm$ 37.47 (-11.4\%) & 529.80 (-6.4\%) & 1.06 (-28.6\%) & 645.36 \\
 & $+$All & 4,727.12 $\pm$ 502.73 (-37.6\%) & 12,696.72 (-18.3\%) & 258.03 $\pm$ 40.39 (-18.5\%) & 507.08 (-10.4\%) & 0.82 (-44.4\%) & 1,184.97 \\
\midrule
\multirow{7}{*}{\shortstack{\sTwoW-M \\ +32}}
 & Baseline   & 17,662.92 $\pm$ 1142.88 & 31,728.04 & 648.46 $\pm$ 53.29 & 1,088.00 & 5.37 & 21.74 \\
 & \framework & 13,920.84 $\pm$ 913.63 (-21.2\%) & 27,985.96 (-11.8\%) & 515.11 $\pm$ 41.02 (-20.6\%) & 954.64 (-12.3\%) & 5.00 (-6.8\%) & 15.75 \\
 & $+$\singleAgentTrajReplan & 13,969.24 $\pm$ 1290.72 (-20.9\%) & 28,034.36 (-11.6\%) & 520.55 $\pm$ 47.99 (-19.7\%) & 960.08 (-11.8\%) & 4.98 (-7.2\%) & 1,030.02 \\
 & $+$\switch & 13,170.48 $\pm$ 933.38 (-25.4\%) & 27,235.60 (-14.2\%) & 506.19 $\pm$ 46.94 (-21.9\%) & 945.72 (-13.1\%) & 4.66 (-13.1\%) & 662.34 \\
 & $+$\trajReplan & 12,728.36 $\pm$ 889.94 (-27.9\%) & 26,793.48 (-15.6\%) & 483.70 $\pm$ 38.19 (-25.4\%) & 923.24 (-15.1\%) & 4.74 (-11.7\%) & 1,331.03 \\
 & $+$\switch $+$\trajReplan & 12,575.88 $\pm$ 927.18 (-28.8\%) & 26,641.00 (-16.0\%) & 506.06 $\pm$ 46.94 (-22.0\%) & 945.60 (-13.1\%) & 4.50 (-16.2\%) & 2,246.07 \\
 & $+$All & 10,855.56 $\pm$ 935.58 (-38.5\%) & 24,920.68 (-21.5\%) & 445.86 $\pm$ 59.13 (-31.2\%) & 885.40 (-18.6\%) & 3.84 (-28.5\%) & 3,654.47 \\
\midrule
\multirow{7}{*}{\shortstack{\dAndE-M \\ +32}}
 & Baseline   & 14,504.16 $\pm$ 1093.14 & 29,079.60 & 547.56 $\pm$ 56.85 & 1,003.04 & 3.13 & 24.64 \\
 & \framework & 11,820.40 $\pm$ 814.17 (-18.5\%) & 26,395.84 (-9.2\%) & 465.76 $\pm$ 51.70 (-14.9\%) & 921.24 (-8.2\%) & 2.84 (-9.1\%) & 22.32 \\
 & $+$\singleAgentTrajReplan & 11,767.72 $\pm$ 982.87 (-18.9\%) & 26,343.16 (-9.4\%) & 457.76 $\pm$ 52.95 (-16.4\%) & 913.24 (-9.0\%) & 2.84 (-9.4\%) & 1,967.62 \\
 & $+$\switch & 11,486.04 $\pm$ 925.09 (-20.8\%) & 26,061.48 (-10.4\%) & 462.12 $\pm$ 53.63 (-15.6\%) & 917.60 (-8.5\%) & 2.64 (-15.7\%) & 868.22 \\
 & $+$\trajReplan & 10,393.32 $\pm$ 893.26 (-28.3\%) & 24,968.76 (-14.1\%) & 431.68 $\pm$ 41.74 (-21.2\%) & 887.16 (-11.6\%) & 2.44 (-22.1\%) & 1,793.65 \\
 & $+$\switch $+$\trajReplan & 10,394.60 $\pm$ 810.37 (-28.3\%) & 24,970.04 (-14.1\%) & 435.08 $\pm$ 53.63 (-20.5\%) & 890.56 (-11.2\%) & 2.34 (-25.2\%) & 2,830.13 \\
 & $+$All & 8,628.84 $\pm$ 782.30 (-40.5\%) & 23,204.28 (-20.2\%) & 378.88 $\pm$ 66.85 (-30.8\%) & 834.36 (-16.8\%) & 1.95 (-37.8\%) & 5,235.96 \\
\midrule
\multirow{2}{*}{\shortstack{\rTwoR \\ +100}}
 & Baseline & 72,018.48 $\pm$ 2951.52 & 134,825.72 & 915.57 $\pm$ 44.57 & 1,543.64 & 3.48 & 341.43 \\
 & CREST    & 54,331.84 $\pm$ 2313.87 (-24.6\%) & 117,139.08 (-13.1\%) & 798.77 $\pm$ 90.45 (-12.8\%) & 1,426.84 (-7.6\%) & 3.09 (-11.3\%) & 347.89 \\
\midrule
\multirow{2}{*}{\shortstack{\sTwoW-L \\ +100}}
 & Baseline & 461,249.72 $\pm$ 14268.73 & 745,418.36 & 4,760.87 $\pm$ 158.15 & 7,602.56 & 16.29 & 6,258.26 \\
 & CREST    & 347,397.56 $\pm$ 7376.90 (-24.7\%) & 631,566.20 (-15.3\%) & 3,640.87 $\pm$ 75.61 (-23.5\%) & 6,482.56 (-14.7\%) & 15.27 (-6.3\%) & 8,352.40 \\
\midrule
\multirow{2}{*}{\shortstack{\dAndE-L \\ +100}}
 & Baseline & 334,300.36 $\pm$ 9998.63 & 592,910.88 & 3,466.45 $\pm$ 99.92 & 6,052.56 & 18.01 & 4,899.67 \\
 & CREST    & 280,740.56 $\pm$ 9483.68 (-16.0\%) & 539,351.08 (-9.0\%) & 2,949.21 $\pm$ 87.30 (-14.9\%) & 5,535.32 (-8.5\%) & 16.69 (-7.3\%) & 8,634.67 \\
\bottomrule
\end{tabular}
\end{adjustbox}
\end{table*}

\begin{table*}[t!]
\caption{Results for the baseline and all CREST variants \textbf{with overhead} ($\Delta=1$). Values are reported as mean $\pm$ standard deviation over 25 instances for normalized metrics (Norm. Cost and Norm. Mksp); other metrics are reported as mean. Percentages denote reductions relative to the baseline. `$N$' denotes the number of agents in the setting.}
\vspace{5pt}
\label{table-main-result-with-overhead}
\centering
\begin{adjustbox}{max width=\textwidth}
\begin{tabular}{c|c|c|c|c|c|c|c}
\toprule
\multicolumn{1}{c|}{\textbf{Map + N}} &
\multicolumn{1}{c|}{\textbf{Method}} &
\multicolumn{1}{c|}{\textbf{Norm. Cost}} &
\multicolumn{1}{c|}{\textbf{Cost}} &
\multicolumn{1}{c|}{\textbf{Norm. Mksp}} &
\multicolumn{1}{c|}{\textbf{Mksp}} &
\multicolumn{1}{c|}{\textbf{\# Switch/Shelf}} &
\multicolumn{1}{c}{\textbf{Time (s)}} \\
\midrule
\multirow{7}{*}{\shortstack{\rTwoR-M \\ +32}}
 & Baseline   & 10,836.56 $\pm$ 833.35 & 18,806.16 & 424.67 $\pm$ 40.45 & 673.72 & 1.50 & 9.11 \\
 & \framework & 8,229.04 $\pm$ 599.76 (-24.1\%) & 16,198.64 (-13.9\%) & 379.47 $\pm$ 58.84 (-10.6\%) & 628.52 (-6.7\%) & 1.24 (-17.0\%) & 7.05 \\
 & $+$\singleAgentTrajReplan & 8,290.64 $\pm$ 681.77 (-23.5\%) & 16,260.24 (-13.5\%) & 378.35 $\pm$ 54.52 (-10.9\%) & 627.40 (-6.9\%) & 1.25 (-16.6\%) & 165.35 \\
 & $+$\switch & 7,852.60 $\pm$ 650.48 (-27.5\%) & 15,822.20 (-15.9\%) & 368.23 $\pm$ 52.76 (-13.3\%) & 617.28 (-8.4\%) & 1.13 (-24.4\%) & 181.34 \\
 & $+$\trajReplan & 7,597.36 $\pm$ 694.61 (-29.9\%) & 15,566.96 (-17.2\%) & 383.87 $\pm$ 50.66 (-9.6\%) & 632.92 (-6.1\%) & 1.06 (-29.2\%) & 458.92 \\
 & $+$\switch $+$\trajReplan & 7,581.92 $\pm$ 688.80 (-30.0\%) & 15,551.52 (-17.3\%) & 376.27 $\pm$ 52.76 (-11.4\%) & 625.32 (-7.2\%) & 1.03 (-31.1\%) & 655.86 \\
 & $+$All & 6,840.92 $\pm$ 711.67 (-36.9\%) & 14,810.52 (-21.2\%) & 338.35 $\pm$ 68.67 (-20.3\%) & 587.40 (-12.8\%) & 0.85 (-43.4\%) & 745.31 \\
\midrule
\multirow{7}{*}{\shortstack{\sTwoW-M \\ +32}}
 & Baseline   & 25,309.88 $\pm$ 1038.67 & 39,375.00 & 889.62 $\pm$ 72.26 & 1,329.16 & 5.44 & 21.71 \\
 & \framework & 18,986.84 $\pm$ 704.88 (-25.0\%) & 33,051.96 (-16.1\%) & 687.90 $\pm$ 68.88 (-22.7\%) & 1,127.44 (-15.2\%) & 4.90 (-10.0\%) & 15.28 \\
 & $+$\singleAgentTrajReplan & 18,870.00 $\pm$ 989.33 (-25.4\%) & 32,935.12 (-16.4\%) & 692.87 $\pm$ 73.41 (-22.1\%) & 1,132.40 (-14.8\%) & 4.88 (-10.4\%) & 526.92 \\
 & $+$\switch & 17,815.88 $\pm$ 708.96 (-29.6\%) & 31,881.00 (-19.0\%) & 677.70 $\pm$ 71.20 (-23.8\%) & 1,117.24 (-15.9\%) & 4.54 (-16.6\%) & 647.26 \\
 & $+$\trajReplan & 17,559.92 $\pm$ 1133.44 (-30.6\%) & 31,625.04 (-19.7\%) & 671.42 $\pm$ 71.18 (-24.5\%) & 1,110.96 (-16.4\%) & 4.63 (-14.8\%) & 1,338.30 \\
 & $+$\switch $+$\trajReplan & 17,227.32 $\pm$ 714.49 (-31.9\%) & 31,292.44 (-20.5\%) & 661.58 $\pm$ 71.20 (-25.6\%) & 1,101.12 (-17.2\%) & 4.41 (-18.8\%) & 2,089.58 \\
 & $+$All & 15,189.16 $\pm$ 885.72 (-40.0\%) & 29,254.28 (-25.7\%) & 593.78 $\pm$ 101.31 (-33.3\%) & 1,033.32 (-22.3\%) & 3.87 (-28.8\%) & 2,459.48 \\
\midrule
\multirow{7}{*}{\shortstack{\dAndE-M \\ +32}}
 & Baseline   & 22,434.04 $\pm$ 1370.29 & 37,009.48 & 805.00 $\pm$ 68.76 & 1,260.48 & 3.21 & 21.33 \\
 & \framework & 18,101.60 $\pm$ 1163.75 (-19.3\%) & 32,677.04 (-11.7\%) & 673.64 $\pm$ 74.85 (-16.3\%) & 1,129.12 (-10.4\%) & 2.77 (-13.7\%) & 22.41 \\
 & $+$\singleAgentTrajReplan & 17,977.44 $\pm$ 1230.08 (-19.9\%) & 32,552.88 (-12.0\%) & 661.40 $\pm$ 73.90 (-17.8\%) & 1,116.88 (-11.4\%) & 2.76 (-14.1\%) & 895.77 \\
 & $+$\switch & 17,148.32 $\pm$ 1333.79 (-23.6\%) & 31,723.76 (-14.3\%) & 648.92 $\pm$ 76.31 (-19.4\%) & 1,104.40 (-12.4\%) & 2.58 (-19.6\%) & 846.90 \\
 & $+$\trajReplan & 16,328.04 $\pm$ 1368.70 (-27.2\%) & 30,903.48 (-16.5\%) & 635.80 $\pm$ 58.85 (-21.0\%) & 1,091.28 (-13.4\%) & 2.44 (-24.0\%) & 1,728.29 \\
 & $+$\switch $+$\trajReplan & 15,974.60 $\pm$ 1120.05 (-28.8\%) & 30,550.04 (-17.5\%) & 647.20 $\pm$ 76.31 (-19.6\%) & 1,102.68 (-12.5\%) & 2.30 (-28.5\%) & 2,514.40 \\
 & $+$All & 14,009.80 $\pm$ 1201.81 (-37.6\%) & 28,585.24 (-22.8\%) & 570.00 $\pm$ 80.09 (-29.2\%) & 1,025.48 (-18.6\%) & 2.00 (-37.7\%) & 3,146.83 \\
\midrule
\multirow{2}{*}{\shortstack{\rTwoR \\ +100}}
 & Baseline & 95,540.28 $\pm$ 4706.80 & 158,347.52 & 1,157.57 $\pm$ 55.04 & 1,785.64 & 3.49 & 327.26 \\
 & CREST    & 70,321.96 $\pm$ 3979.80 (-26.4\%) & 133,129.20 (-15.9\%) & 967.41 $\pm$ 138.13 (-16.4\%) & 1,595.48 (-10.6\%) & 3.02 (-13.3\%) & 325.86 \\
\midrule
\multirow{2}{*}{\shortstack{\sTwoW-L \\ +100}}
 & Baseline & 591,783.24 $\pm$ 12374.68 & 875,951.88 & 6,061.63 $\pm$ 254.52 & 8,903.32 & 16.44 & 5,541.09 \\
 & CREST    & 441,672.72 $\pm$ 8415.37 (-25.4\%) & 725,841.36 (-17.1\%) & 4,605.95 $\pm$ 125.08 (-24.0\%) & 7,447.64 (-16.3\%) & 15.08 (-8.2\%) & 9,019.23 \\
\midrule
\multirow{2}{*}{\shortstack{\dAndE-L \\ +100}}
 & Baseline & 488,965.80 $\pm$ 9204.92 & 747,576.32 & 5,021.25 $\pm$ 122.47 & 7,607.36 & 18.34 & 4,686.87 \\
 & CREST    & 393,843.16 $\pm$ 9617.43 (-19.5\%) & 652,453.68 (-12.7\%) & 4,099.37 $\pm$ 249.80 (-18.4\%) & 6,685.48 (-12.1\%) & 16.56 (-9.7\%) & 9,459.79 \\
\bottomrule
\end{tabular}
\end{adjustbox}
\end{table*}

\end{document}